\documentclass[runningheads]{llncs}
\usepackage{amsmath,amssymb} 
\usepackage{color,soul}
\usepackage{multirow}
\usepackage{tabularx}
\usepackage[tableposition=top]{caption}
\usepackage{tikz}
\usetikzlibrary{shapes,matrix,arrows,decorations.pathmorphing, positioning, calc, decorations.markings}
\usepackage{graphicx}
\usepackage{subfigure}
\usepackage{stackengine}
\usepackage[pagebackref=false,breaklinks=true,letterpaper=true,colorlinks,bookmarks=false]{hyperref}

\newcommand{\etal}{\textit{et al}.}
\newcommand{\ie}{\textit{i}.\textit{e}.}
\newcommand{\eg}{\textit{e}.\textit{g}.}

\begin{document}

\newcommand{\point}{
    \raise0.7ex\hbox{.}
    }


\pagestyle{headings}

\mainmatter

\title{Deep Relative Attributes} 

\titlerunning{Deep Relative Attributes} 

\authorrunning{Yaser Souri, Erfan Noury, Ehsan Adeli} 

\author{Yaser Souri$^1$, Erfan Noury$^2$, Ehsan Adeli$^3$} 
\institute{$^1$Sobhe \quad $^2$Sharif University of Technology \\ $^3$University of North Carolina at Chapel Hill} 



\maketitle

\begin{abstract}
Visual attributes are great means of describing images or scenes, in a way both humans and computers understand. In order to establish a correspondence between images and to be able to compare the strength of each property between images, relative attributes were introduced. However, since their introduction, hand-crafted and engineered features were used to learn increasingly complex models for the problem of relative attributes. This limits the applicability of those methods for more realistic cases. We introduce a deep neural network architecture for the task of relative attribute prediction. A convolutional neural network (ConvNet) is adopted to learn the features by including an additional layer (ranking layer) that learns to rank the images based on these features. We adopt an appropriate ranking loss to train the whole network in an end-to-end fashion. Our proposed method outperforms the baseline and state-of-the-art methods in relative attribute prediction on various coarse and fine-grained datasets. Our qualitative results along with the visualization of the saliency maps show that the network is able to learn effective features for each specific attribute. Source code of the proposed method is available at \href{https://github.com/yassersouri/ghiaseddin}{https://github.com/yassersouri/ghiaseddin}.
\end{abstract}

\section{Introduction}

Visual attributes are linguistic terms that bear semantic properties of (visual) entities, often shared among categories. They are both human understandable and machine detectable, which makes them appropriate for better human machine communications. Visual attributes have been successfully used for many applications, such as image search \cite{whittlesearch}, interactive fine-grained recognition, \cite{branson13,branson10} and zero-shot learning \cite{6571196,parikh2011}. 

\begin{figure}
\centering
\scalebox{.3}
{
    \begin{tikzpicture}
        \node (p1A) {\includegraphics[scale=0.65]{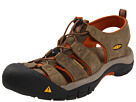}};
        \node (p1op) [right=0.0cm of p1A, scale=3] {$>$};
        \node (p1B) [right=-0.4cm of p1op] {\includegraphics[scale=0.65]{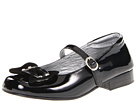}};
        \node (training) [left=1cm of p1A, align=center, scale=4] {Training \\ set};
        \draw[line width=2, rounded corners]($(p1A.north west)+(-0.0,0.5)$)  rectangle ($(p1B.south east)+(0.0,-0.5)$);
        
        \node (p2B) [right=0.0cm of p1B] {\includegraphics[scale=0.65] {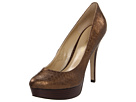}};
        \node (p2op) [right=0.0cm of p2B, scale=3] {$<$};
        \node (p2A) [right=-0.4cm of p2op] {\includegraphics[scale=0.65] {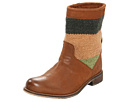}};
        \draw[line width=2, rounded corners] ($(p2B.north west)+(0.2,0.5)$) rectangle ($(p2A.south east)+(0.0, -0.5)$);

        \node (dots) [right=0.1cm of p2A, scale=4] {\dots};

        \node (p3A) [right=0.0cm of dots] {\includegraphics[scale=0.65]{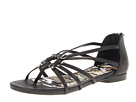}};
        \node (p3op) [right=0.0cm of p3A, scale=3] {$<$};
        \node (p3B) [right=-0.4cm of p3op] {\includegraphics[scale=0.65]{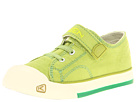}};
        \draw[line width=2, rounded corners] ($(p3A.north west)+(-0.2,0.5)$) rectangle ($(p3B.south east)+(0.2, -0.5)$);

        \draw[line width=3, rounded corners=15pt, red!30] ($(p1A.north west)+(-0.4, 1)$) rectangle ($(p3B.south east)+(0.6, -1)$);

        \path (p1A.south east) -- (p3B.south west) node (mid) [below=1cm, pos=0.5] {};

        \node (uop) [below=4cm of mid, scale=4] {$?$};
        \node (puA) [left=0.5cm of uop] {\includegraphics[scale=1]{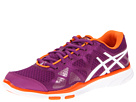}};
        \node (puB) [right=0.5cm of uop] {\includegraphics[scale=1]{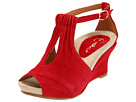}};
        \draw [line width=1, rounded corners=0.75cm] ($(puA.north west)+(-0.2, 0.5)$) rectangle ($(puB.south east)+(0.2, -0.5)$);
        \node (test) [left=1cm of puA, align=center, scale=4] {Test \\ instance};
        
        \draw [->, line width=10] (mid.south) -- ($(uop.north) + (0cm, 2cm)$);
    \end{tikzpicture}
}
\caption{Visual Relative Attributes. This figure shows samples of training pairs of images from the UT-Zap50K dataset, comparing shoes in terms of the \textit{comfort} attribute (top). The goal is to compare a pair of two novel images of shoes, respective to the same attribute (bottom).}
\label{fig.1}
\end{figure}
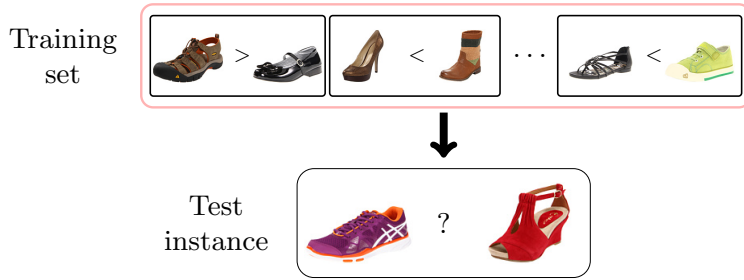

Traditionally, visual attributes were treated as binary concepts \cite{ferrari2007learning,Farhadi09describingobjects}, as if they are present or not, in an image. Parikh and Grauman~\cite{parikh2011} introduced a more natural view on visual attributes, in which pairs of visual entities can be compared, with respect to their relative strength of any specific attribute. With a set of human assessed relative orderings of image pairs, they learn a global ranking function for each attribute that can be used to compare a pair of two novel images respective to the same attribute (Figure \ref{fig.1}).
While binary visual attributes relate properties to entities (\eg, a dog being furry), relative attributes make it possible to relate entities to each other in terms of their properties (\eg, a bunny being furrier than a dog).

Many have tried to build on the seminal work of Parikh and Grauman \cite{parikh2011} with more complex and task-specific models for ranking, while still using hand-crafted visual features, such as GIST \cite{Aude01} and HOG \cite{hog}. Recently, Convolutional Neural Networks (ConvNets) have proved to be successful in various visual recognition tasks, such as image classification \cite{Krizhevsky2012ImageNetCW}, object detection \cite{RCNN} and image segmentation \cite{fullyconv}. Many ascribe the success of ConvNets to their ability to learn multiple layers of visual features from the data. 

In this work, we propose to use a ConvNet-based architecture comprising of a feature learning and extraction and ranking portions. This network is used to learn the ranking of images, using relatively annotated pairs of images with similar and/or different strengths of some particular attribute. The network learns a series of visual features, which are known 
 to perform better than the engineered visual features for various tasks \cite{offtheshelf}. These layers could simply be learned through gradient descent. As a result, it would be possible to learn (or fine-tune) the features through back-propagation, while learning the ranking layer.
Interweaving the two processes leads to a set of learned features that appropriately characterizes each single attribute. Our qualitative investigation of the learned feature space further confirms this assumption. This escalates the overall performance and is the main advantage of our proposed method over previous methods. 
Furthermore, our proposed model can effectively utilize pairs of images with equal annotated attribute strength. The equality relation can happen quite frequently when humans are qualitatively deciding about the relations of attributes in images. In previous works, this is often overlooked and mainly inequality relations are exploited. Our proposed method incorporates an easy and elegant way to deal with equality relations (\ie, an attribute is similarly strong in two images). 
In addition, it is noteworthy to pinpoint that by exploiting the saliency maps of the learned features for each attribute, similar to \cite{saliency}, we can discover the pixels which contribute the most towards an attribute in the image. This can be used to coarsely localize the specific attribute.

Our approach achieves very competitive results and improves the state-of-the-art (with a large margin in some datasets) on major publicly available datasets for relative attribute prediction, both coarse and fine-grained, while many of the previous works targeted only one of the two sets of problems (coarse or fine-grained), and designed a method accordingly.

The rest of the paper is organized as follows: Section \ref{sec.2} discusses the related works. Section \ref{sec.3} illustrates our proposed method. Then, Section \ref{sec.4} exhibits the experimental setup and results, and finally, Section \ref{sec.5} concludes the paper.

\section{Related Works}
\label{sec.2}

We usually describe visual concepts with their attributes. 
Attributes are, therefore, mid-level representations for describing objects and scenes. In an early work on attributes, Farhadi \etal~\cite{Farhadi09describingobjects} proposed to describe objects using mid-level attributes. In another work \cite{Farhadi2010EveryPT}, the authors described images based on 
a semantic triple ``object, action, scene". In the recent years, attributes have shown great performance in object recognition \cite{Farhadi09describingobjects,7298613}, action recognition \cite{6838985,5995353} and event detection \cite{6475038}. Lampert \etal~\cite{6571196} predicted unseen objects using a zero-shot learning framework, incorporating the binary attribute representation of the objects. 

Although detection and recognition based on the presence of attributes appeared to be quite interesting, comparing attributes enables us to easily and reliably search through high-level data derived from \eg, documents or images. For instance, Kovashka \etal~\cite{KovashkaG13} proposed a relevance feedback strategy for image search using attributes and their comparisons. In order to establish the capacity for comparing attributes, we need to move from binary attributes towards describing attributes relatively. In the recent years, relative attributes have attracted the attention of many researchers.
For instance, a linear relative comparison function is learned in \cite{parikh2011}, based on RankSVM \cite{Joachims2002} and a non-linear strategy in \cite{Li2012RelativeFF}. In another work, Datta \etal~\cite{5771429} used trained rankers for each facial image feature and formed a global ranking function for attributes.

For the process of learning the attributes, different types of low-level image features are often  incorporated. For instance, Parikh and Grauman~\cite{parikh2011} used 512-dimensional GIST \cite{Aude01} descriptors as image features, while Jayaraman \etal~\cite{6909607} used histograms of image features, and reduced their dimensionality using PCA. Other works tried learning attributes through \eg, local learning \cite{1641014} or fine-grained comparisons \cite{Yu2014}. Yu and Grauman \cite{Yu2014} proposed a local learning-to-rank framework for fine-grained visual comparisons, in which the ranking model is learned using only analogous training comparisons. In another work \cite{Yu2015}, they proposed a local Bayesian model to rank images, which are hardly distinguishable for a given attribute. However, none of these methods leverage the effectiveness of feature learning methods and only use engineered and hand-crafted features for predicting relative attributes. 

As could be inferred from the literature, it is very hard to decide what low-level image features to use for identifying and comparing visual attributes. Recent studies show that features learned through the convolutional neural networks (CNNs) \cite{LeCun1989HandwrittenDR} (also known as deep features) could achieve great performance for image classification \cite{Krizhevsky2012ImageNetCW} and object detection \cite{6909475}. Zhang \etal~\cite{6909608} utilized CNNs for classifying binary attributes. In other works, Escorcia \etal~\cite{Escorcia_2015_CVPR} proposed CCNs with attribute centric nodes within the network for establishing the relationships between visual attributes. Shankar \etal~\cite{Shankar_2015_CVPR} proposed a weakly supervised setting on convolutional neural networks, applied for attribute detection. Khan \etal~\cite{khan15} used deep features for describing human attributes and thereafter for action recognition, and Huang \etal~\cite{Huang_2015_ICCV} used deep features for cross-domain image retrieval based on binary attributes.  

Neural networks have also been extended for learning-to-rank applications. One of the earliest networks for ranking was proposed by Burges\etal~\cite{Burges2005}, known as RankNet. The underlying model in RankNet maps an input feature vector to a Real number. The model is trained  by presenting the network pairs of input training feature vectors with differing labels. Then, based on how they should be ranked, the underlying model parameters are updated. This model is used in different fields for ranking and retrieval applications, \eg, for personalized search \cite{song2014} or content-based image retrieval \cite{Wan2014}. In another work, Yao \etal~\cite{YaoCVPR2016} proposed a ranking framework for videos for first-person
video summarization, through recognizing video highlights. They incorporated both spatial and temporal streams through 2D and 3D CNNs and detect the video highlights.

\section{Proposed Method}
\label{sec.3}

We propose to use a ConvNet-based deep neural network that is trained to optimize an appropriate ranking loss for the task of predicting relative attribute strength. The network architecture consists of two parts, the \textit{feature learning and extraction} part and the \textit{ranking} part.

The feature learning and extraction part takes a fixed size image, $I_i$, as input and outputs the learned feature representation for that image $\psi_i \in \mathbb{R}^d$.
Over the past few years, different network architectures for computer vision problems have been developed. These deep architectures can be used for extracting and learning features for different applications.
For the current work, outputs of an intermediate layer, like the last layer before the probability layer, from a ConvNet architecture (\eg, AlexNet \cite{Krizhevsky2012ImageNetCW}, VGGNet \cite{verydeep} or GoogLeNet \cite{googlenet}) can be incorporated. 
In our experiments we use the VGG-16 architecture \cite{verydeep} with the last fully connected layer (the class probabilities) removed. This architecture takes as input a 224x224 RGB image and consists of 13, 3x3 convolutional layers with max pooling layers in between. In addition, it has 2 fully connected layers on top of the convolutional layers. For details on the architecture see \cite{verydeep}.

One of the most widely used models for relative attributes in the literature is RankSVM \cite{Joachims2002}. However,
in our case, we seek a neural network-based ranking procedure, to which relatively ordered pairs of feature vectors are provided during training. This procedure should learn to map each feature vector to an absolute ranking, for testing purpose. Burges \etal~\cite{Burges2005} introduced such a neural network based ranking procedure that exquisitely fits our needs. 
We adopt a similar strategy and thus, the ranking part of our proposed network architecture is analogous to \cite{Burges2005} (referred to as RankNet).

During training for a minibatch of image pairs and their target orderings, the output of the feature learning and extraction part of the network is fed into the ranking part and a ranking loss is computed. The loss is then back-propagated through the network, which enables us to simultaneously learn the weights of both feature learning and extraction (ConvNet) and ranking (RankNet) parts of the network.  
Further with back-propagation we can calculate the derivative of the estimated ordering with respect to the pixel values.
In this way, we can generate saliency maps for each attribute (see section \ref{sec.4.5}). These saliency maps exhibit interesting properties, as they can be used to localize the regions in the image that are informative about the attribute.

\subsection{RankNet: Learning to Rank Using Gradient Descent}\label{sec3.1}

This section briefly overviews the RankNet 
procedure in our context.
Given a set (of size $n$) of pairs of sample feature vectors $\big\{( \psi_{1}^{(k)}, \psi_{2}^{(k)} ) | k \in \{1, \dots, n\} \big\} \in \mathbb{R}^{d \times d}$, and target probabilities $\big\{ t_{12}^{(k)} | k \in \{1, \dots, n\} \big\}$, which indicate the probability of  
sample $\psi_{1}^{(k)}$ being ranked higher than sample $\psi_{2}^{(k)}$. 
We would like to learn a ranking function $f : \mathbb{R}^d \mapsto \mathbb{R}$, such that $f$ specifies the ranking order of a set of features. Here, $f(\psi_i) > f(\psi_j)$ indicates that the 
feature vector $\psi_i$ is ranked higher than $\psi_j$, denoted by $\psi_i \triangleright \psi_j$. The RankNet model \cite{Burges2005} provides an elegant procedure based on neural networks 
to learn the function $f$ from a set of pairs of samples and target probabilities.

Denoting $r_i \equiv f(\psi_i)$, RankNet models the mapping from rank estimates to posterior probabilities $p_{ij} = P(\psi_i \triangleright \psi_j)$ using a logistic function 

\begin{equation}
p_{ij} := \frac{1}{1 + e^{-(r_i - r_j)}}.
\label{eq1}
\end{equation}

\begin{figure}
    \centering
    \begin{subfigure}
        \centering
        \includegraphics[width=5cm]{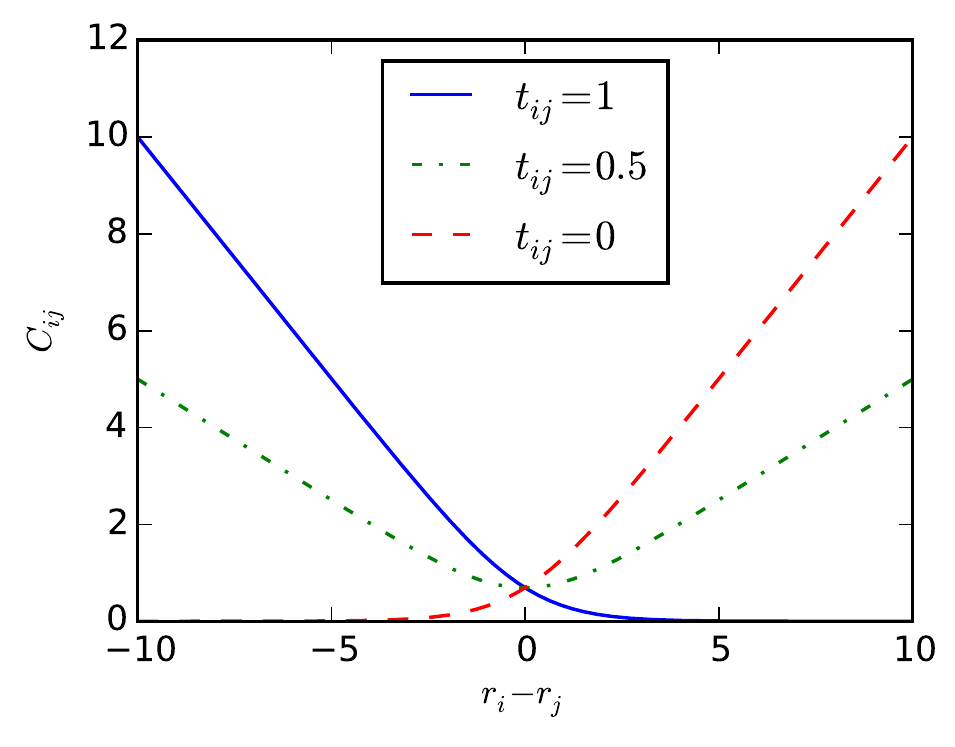}
    \end{subfigure}
    \begin{subfigure}
        \centering
        \includegraphics[width=5cm]{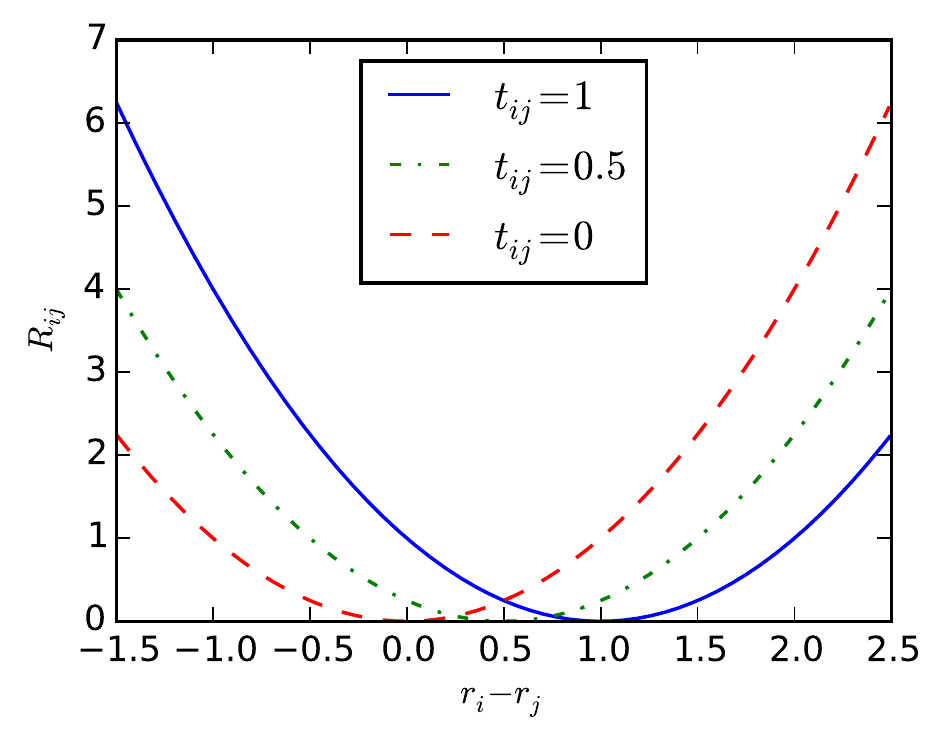}
    \end{subfigure}
    \caption{The ranking loss value for three values of the target probability (left). The squared loss value for three values of the target probability, typically used for regression (right).}
    \label{fig.2}
\end{figure}

The loss for the sample pair of feature vectors $(\psi_i, \psi_j)$ along with target probability $t_{ij}$ is defined as

\begin{equation}
C_{ij} := - t_{ij} \log (p_{ij}) - (1 - t_{ij}) \log (1 - p_{ij}),
\label{eq2}
\end{equation}
which is the binary cross entropy loss.
Figure \ref{fig.2} (left) plots the loss value $C_{ij}$ as a function of $r_i - r_j$ for three values of target probability $t_{ij} \in \{0, 0.5, 1\}$. This function is quite suitable for ranking purposes, as it acts differently compared to regression functions.
Specifically, we are not interested in regression instead of ranking for two reasons: First, we cannot regress the absolute rank of images, since the annotations are only available in pairwise ordering for each attribute, in relative attribute datasets (see section \ref{sec.4.1}). Second, regressing the difference $r_i - r_j$ to $t_{ij}$ is inappropriate. To understand this, let's consider the squared loss

\begin{equation}
R_{ij} = \big[(r_i - r_j) - t_{ij}\big]^2,
\end{equation}
which is typically used for regression, illustrated in Figure \ref{fig.2} (right). We observe that the regression loss forces the difference of rank estimates to be a specific value and disallows over-estimation. Furthermore, its quadratic natures makes it 
sensitive to noise. This sheds light into why regression objective is the wrong objective to optimize when the goal is ranking.

Note that when $t_{ij} = 0.5$, and no information is available about the relative rank of the two samples, the ranking cost becomes symmetric. This can be used as a way to train on patterns that are desired to have similar ranks. This is somewhat not much studied in the previous works on relative attributes.
Furthermore, this model asymptotically converges to a linear function which makes it more appropriate for problems with noisy labels. 

Training this model is possible using stochastic gradient descent or its variants like RMSProp.
While testing, we only need to estimate the value of $f(\psi_i)$, which resembles the absolute rank of the testing sample. Using $f(\psi_i)$s, we can easily infer both absolute or relative ordering of the testing pairs.

\subsection{Deep Relative Attributes}\label{sec3.2}

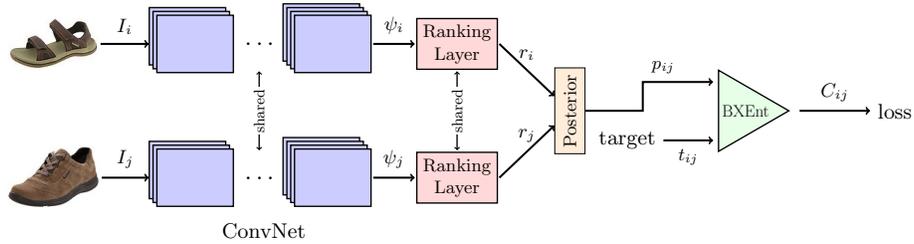
\begin{figure}
\centering
\resizebox{\linewidth}{!}
{
    \begin{tikzpicture}
        \node (im1) at (0cm,1cm)  {\includegraphics[scale=1.3]{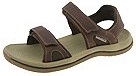}};
        \node (im2) at (0cm, -6cm)  {\includegraphics[scale=1]{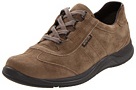}};

        \node (tconv1) at (5.1cm, 1cm) {};
        \draw [fill=blue!20, line width=2pt] (5cm,2.9cm) rectangle (9cm,-0.1cm);
        \draw [fill=blue!20, line width=2pt] (5.2cm,2.7cm) rectangle (9.2cm,-0.3cm);
        \draw [fill=blue!20, line width=2pt] (5.4cm,2.5cm) rectangle (9.4cm,-0.5cm);

        \node (bconv1) at (5.1cm, -6cm) {};
        \draw [fill=blue!20, line width=2pt] (5cm,-4.1cm) rectangle (9cm,-7.1cm);
        \draw [fill=blue!20, line width=2pt] (5.2cm,-4.3cm) rectangle (9.2cm,-7.3cm);
        \draw [fill=blue!20, line width=2pt] (5.4cm,-4.5cm) rectangle (9.4cm,-7.5cm);

        \path [draw, ->, line width=3] (im1.east) -- node [above, scale=3.2] {$I_i$} (tconv1);
        \path [draw, ->, line width=3] (im2.east) -- node [above, scale=3.2] {$I_j$} (bconv1);

        \node (tconv2) at (12cm, 1cm) {};
        \draw [fill=blue!20, line width=2pt] (12cm,3.1cm) rectangle (16cm,0.1cm);
        \draw [fill=blue!20, line width=2pt] (12.2cm,2.9cm) rectangle (16.2cm,-0.1cm);
        \draw [fill=blue!20, line width=2pt] (12.4cm,2.7cm) rectangle (16.4cm,-0.3cm);
        \draw [fill=blue!20, line width=2pt] (12.6cm,2.5cm) rectangle (16.6cm,-0.5cm);

        \node (bconv2) at (12cm, -6cm) {};
        \draw [fill=blue!20, line width=2pt] (12cm,-3.9cm) rectangle (16cm,-6.9cm);
        \draw [fill=blue!20, line width=2pt] (12.2cm,-4.1cm) rectangle (16.2cm,-7.1cm);
        \draw [fill=blue!20, line width=2pt] (12.4cm,-4.3cm) rectangle (16.4cm,-7.3cm);
        \draw [fill=blue!20, line width=2pt] (12.6cm,-4.5cm) rectangle (16.6cm,-7.5cm);

        \path (tconv1.east)+(4.65cm, 0cm) -- node [scale=4]{\dots} (tconv2);
        \path (bconv1.east)+(4.65cm, 0cm) -- node [scale=4]{\dots} (bconv2);

        \node (convnet) [below, scale=3.5] at (11cm, -8cm) {ConvNet};

        \node (tconvout) at (16.6cm, 1cm) {};
        \node (trank) at (21.1cm, 1cm) [rectangle, draw, fill=red!15, scale=3, align=center, line width=2pt] {Ranking \\  Layer};
        \path [draw, line width=3, ->] (tconvout) -- node[above, scale=3.2] {$\psi_i$} (trank.west);

        \node (bconvout) at (16.6cm, -6cm) {};
        \node (brank) at (21.1cm, -6cm) [rectangle, draw, fill=red!15, scale=3, align=center, line width=2pt] {Ranking \\  Layer};
        \path [draw, line width=3, ->] (bconvout) -- node[above, scale=3.2] {$\psi_j$} (brank.west);

        \node (dashcenter) at (10.75cm, 0cm) {};
        \path [draw, line width=2, <->] (trank.south) -- node [scale=2.5, rectangle, fill=white, line width=0, inner sep=1pt] {\rotatebox{90}{shared}} (brank.north);
        \path [draw, line width=2, <->] (trank.south -| dashcenter) -- node [scale=2.5, rectangle, fill=white, line width=0, inner sep=1pt] {\rotatebox{90}{shared}} (brank.north -| dashcenter);

        \node (posterior) at (27cm, -2.5cm) [rectangle, draw, fill=orange!15, scale=3, minimum width=15] {\rotatebox{90}{Posterior}};

        \path [draw, line width=3, ->] (trank.east) -- node [scale=3.2, above] {$r_i$} (posterior);
        \path [draw, line width=3, ->] (brank.east) -- node [scale=3.2, above] {$r_j$} (posterior);

        \node (bxent) at (36cm, -2.5cm) [draw,fill=green!10, regular polygon, regular polygon sides=3, shape border rotate=-90, scale=8, line width=2] {};
        \node at (36.2cm, -2.5cm) {\Huge BXEnt};

        \node [align=center, minimum height=3mm] (target) at (30cm , -4.05cm) [scale=3.5] {target};

        \draw [line width=3, ->] (posterior.east) -- ++(3cm, 0cm) |- node [above right, scale=3] {$p_{ij}$}  (bxent.north west);
        \draw [line width=3, ->] (target.east) -- node [below, scale=3] {$t_{ij}$}  (bxent.south west |- target.east);

        \node (loss) [scale=3.5] at (44cm, -2.5cm) {loss};

        \draw [line width=3, ->] (bxent.east) -- node [above, scale=3.2] {$C_{ij}$}  (loss.west);
    \end{tikzpicture}
}
\caption{The overall schematic view of the proposed method during training. The network consists of two parts, the \textit{feature learning and extraction} part (labeled ConvNet in the figure), and the \textit{ranking} part (the Ranking Layer). Pairs of images are presented to the network with their corresponding target probabilities. This is used to calculate the loss, which is then back-propagated through the network to update the weights.}
\label{fig.3}
\end{figure}

Our proposed model is depicted in figure \ref{fig.3}. The model is trained separately, for each attribute. During training, pairs of images $(I_i, I_j)$ are presented to the network, together with the target probability $t_{ij}$. If for the attribute of interest $I_i \triangleright I_j$ (image $i$ exhibits more of the attribute than image $j$), then $t_{ij}$ is expected to be larger than $0.5$ depending on our confidence on the relative ordering of $I_i$ and $I_j$. Similarly, if $I_i \triangleleft I_j$, then $t_{ij}$ is expected to be smaller than $0.5$, and if it is desired that the two images have the same rank, $t_{ij}$ is expected to be $0.5$. Because of the nature of the datasets, we chose $t_{ij}$ from the set $\{0, 0.5, 1 \}$, according to the available annotations in the dataset.

The pair of images then go though the feature learning and extraction part of the network (ConvNet). This procedure maps the images 
onto feature vectors $\psi_i$ and $\psi_j$, respectively. Afterwards, these feature vectors go through the ranking layer, as described in section \ref{sec3.1}. We choose the ranking layer to be a fully connected neural network layer with linear activation function, a single output neuron and weights $w$ and $b$. It maps the feature vector $\psi_i$ to the estimated absolute rank of that feature vector, $r_i \in \mathbb{R}$, where

\begin{equation}
r_i := w^T \psi_i + b.
\end{equation}

The two estimated ranks $r_i$ and $r_j$, for the two images $I_i$ and $I_j$ in comparison, are then combined (using Equation \eqref{eq1}) to output the estimated posterior probability $p_{ij} = P(I_i \triangleright I_j)$. 
This estimated posterior probability is used along with the target probability $t_{ij}$ to calculate the loss, as in Equation \eqref{eq2}. This loss is then back-propagated through the network and is used to update the weights of the whole network, including both the weights of the feature learning and extraction sub-network and the ranking layer.

\begin{figure}
\centering
\resizebox{\linewidth}{!}
{
    \begin{tikzpicture}
        \node (im1) at (0cm,1cm) {\includegraphics[scale=1]{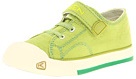}};

        \node (tconv1) at (5.1cm, 1cm) {};
        \draw [fill=blue!20] (5cm,2.9cm) rectangle (9cm,-0.1cm);
        \draw [fill=blue!20] (5.2cm,2.7cm) rectangle (9.2cm,-0.3cm);
        \draw [fill=blue!20] (5.4cm,2.5cm) rectangle (9.4cm,-0.5cm);

        \path [draw, ->, line width=3] (im1.east) -- node [above, scale=3] {$I_k$} (tconv1);

        \node (tconv2) at (12cm, 1cm) {};
        \draw [fill=blue!20] (12cm,3.1cm) rectangle (16cm,0.1cm);
        \draw [fill=blue!20] (12.2cm,2.9cm) rectangle (16.2cm,-0.1cm);
        \draw [fill=blue!20] (12.4cm,2.7cm) rectangle (16.4cm,-0.3cm);
        \draw [fill=blue!20] (12.6cm,2.5cm) rectangle (16.6cm,-0.5cm);

        \path (tconv1.east)+(4.65cm, 0cm) -- node [scale=4]{\dots} (tconv2);

        \node (tconvout) at (16.6cm, 1cm) {};
        \node (trank) at (21.5cm, 1cm) [rectangle, draw, fill=red!20, scale=3, align=center] {Ranking \\  Layer};
        \path [draw, line width=3, ->] (tconvout) -- node[above, scale=3] {$\psi_k$} (trank.west);
        
        \node [scale=3] (posterior) at (26.5cm, 1cm) {$r_k$};

        \path [draw, line width=3, ->] (trank.east) -- (posterior);
    \end{tikzpicture}
}
\caption{During testing, we only need to evaluate $r_k$ for each testing image. Using this value, we can infer the relative or absolute ordering of testing images, for the attribute of interest.}
\label{fig.4}
\end{figure}
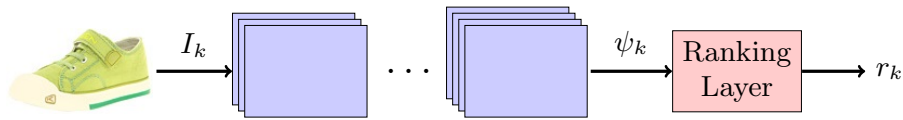

During testing (Figure \ref{fig.4}), we need to calculate the estimated absolute rank $r_k$ for each testing image $I_k$. Using these estimated absolute ranks, we can then easily infer both the relative or absolute attribute ordering, for all testing pairs.

\section{Experiments}\label{sec.4}

To evaluate our proposed method, we quantitatively compare it  with the state-of-the-art methods, as well as an informative baseline on all publicly available benchmarks for relative attributes to our knowledge. Furthermore, we perform multiple qualitative experiments to demonstrate the capability and superiority of our method.

\subsection{Datasets}\label{sec.4.1}

To assess the performance of the proposed method, we have evaluated it on all publicly available datasets to our knowledge: \textbf{Zappos50K} \cite{Yu2014} (both coarse and fine-grained versions), \textbf{LFW-10} \cite{Sandeep_2014_CVPR} and for the sake of completeness and comparison with previous works, on \textbf{PubFig} and \textbf{OSR} datasets of \cite{parikh2011}.

\textbf{UT-Zap50K} \cite{Yu2014} dataset is a collection of images with annotations for relative comparison of 4 attributes. This dataset contains two collections: Zappos50K-1, in which relative attributes are annotated for coarse pairs, where the comparisons are relatively easy to interpret, and Zappos50K-2, where relative attributes are annotated for fine-grained pairs, for which making the distinction between them is hard according to human annotators.
Training set for Zappos50K-1 contains approximately 1500 to 1800 annotated pairs of images for each attribute. These are divided into 10 train/test splits which are provided alongside the dataset and used in this work. Meanwhile, Zappos50K-2 only contains a test set of approximately 4300 pairs, while its training set is the combination of training and testing sets of Zappos50K-1.

We have also conducted experiments on the \textbf{LFW-10} \cite{Sandeep_2014_CVPR} dataset. This dataset has 2000 images of faces of people and annotations for 10 attributes. For each attribute, a random subset of 500 pairs of images have been annotated for each training and testing set.

\textbf{PubFig} \cite{parikh2011} dataset (a set of public figure faces), consists of 800 facial images of 8 random subjects, with 11 attributes.
\textbf{OSR} \cite{parikh2011} dataset contains 2688 images of outdoor scenes in 8 categories, for which 6 relative attributes are defined.
The ordering of samples in both PubFig and OSR datasets are annotated in a category level, \ie, all images in a specific category may be ranked higher, equal, or lower than all images in another category, with respect to an attribute. This sometimes causes annotation inconsistencies \cite{Sandeep_2014_CVPR}.
In our experiments, we have used the provided training/testing split of PubFig and OSR datasets.

\subsection{Experimental setup}
We train our proposed model (described in Section \ref{sec.3}) for each attribute, separately. In our proposed model, it is possible  to train multiple attributes at the same time, however, this is not done due to the structure of the datasets, in which for each training pair of images only a certain attribute is annotated.

We have used the Lasagne \cite{lasagne} deep learning framework to implement our model. 
In all our experiments, for the feature learning and extraction part of the network,
we use the VGG-16 model of \cite{verydeep} and trim out the probability layer (all layers up to fc7 are used, only the probability layer is not included).
We initialize the weights of the model using a pretrained model on ILSVRC 2014 dataset \cite{ilsvrc2014} for the task of image classification. These weights are fine-tuned as the network learns to predict the relative attributes (see section \ref{sec.qres}). The weights $w$ of the ranking layer are initialized using the Xavier method \cite{glorot}, and the bias is initialized to 0.

For training, we use stochastic gradient descent with RMSProp \cite{Tieleman2012} updates and minibatches of size 32 (16 pair of images).
We set the learning rate of the feature learning and extraction layers of the network to $10^{-5}$ and the ranking layer to $10^{-4}$ for all experiments initially, then RMSProp changes the learning rates dynamically during training. We have also used weight decay ($\ell_2$ norm regularization), with a fixed $10^{-5}$ multiplier. Furthermore, when calculating the binary cross entropy loss, we clip the estimated posterior $p_{ij}$ to be in the range $[10^{-7}, 1 - 10^{-7}]$. This is used to prevent the loss from diverging.

In each epoch, we randomly shuffle the training pairs. The number of epochs of training were chosen to reflect the training size. For Zappos50K and LFW-10 datasets, we train for 25 and 40 epochs, respectively. For PubFig and OSR datasets, we train for 2 epochs due to the large number of training sample pairs. When performing evaluation on OSR the total number of pairs is too large (around 3 million pairs) we only evaluate on a 5\% random subset of them. 

\subsection{Baseline}

As a baseline, we have also included results for the RankSVM method (as in \cite{parikh2011}), when the features given to the method were computed from the output of the VGG-16 pretrained network on ILSVRC 2014. 

Using this baseline we can evaluate the extent of effectiveness of off-the-shelf ConvNet features \cite{offtheshelf}  for the task of ranking. In a sense, comparing this baseline with our proposed method reveals the effect of features fine-tuning, for the task. 

\subsection{Quantitative Results}

Following \cite{parikh2011,Yu2014,Sandeep_2014_CVPR}, we report the accuracy in terms of the percentage of correctly ordered pairs. For our proposed method, we report the mean accuracy and standard deviation over 3 separate runs. 

Table \ref{tab:osr} and \ref{tab:pubfig} shows our results on the OSR and PubFig dataset respectively. Our method outperforms the baseline and the state-of-the-art on this dataset by a considerable margin, on most attributes. These are relatively easy datasets but have their own challenges. Specifically the OSR dataset contains attributes like "Perspective" which are very generic, high level and global in the image, which might not correspond easily to local low level image features.
We think that our proposed method is specially well suited for such cases. 

\begin{table*}[t!]
\caption{Results for the OSR dataset}
\centering
\resizebox{\textwidth}{!}{
\begin{tabular}{l| c | c | c | c | c | c | c }
\textbf{Method} & \textbf{Natural} & \textbf{Open} &  \textbf{Perspective} & \textbf{Large Size} & \textbf{Diag} & \textbf{ClsDepth} & \textbf{Mean}\\ \hline
 Relative Attributes~\cite{parikh2011} &  95.03 & 90.77 & 86.73 & 86.23 & 86.50 & 87.53 & 88.80 \\
 Relative Forest~\cite{Li2012RelativeFF} & 95.24 & 92.39 & 87.58 & 88.34 & 89.34 & 89.54 & 90.41 \\
 Fine-grained Comparison~\cite{Yu2014} & 95.70 & 94.10 & 90.43 & 91.10 & 92.43 & 90.47 & 92.37 \\
 VGG16-fc7 (baseline) & 98.00 & 94.46 & 92.92 & 94.08 & 94.91 & 95.02 & 94.90 \\
 \hline
 \multirow{2}{*}{RankNet (ours)} & 99.40  & 97.44  & 96.88  & 96.79  & 98.43  & 97.65  & \textbf{97.77}  \\
        & \scriptsize{($\pm$ 0.10)} & \scriptsize{($\pm$ 0.16)} & \scriptsize{($\pm$ 0.13)} & \scriptsize{($\pm$ 0.32)} & \scriptsize{($\pm$ 0.23)} & \scriptsize{($\pm$ 0.16)} & \scriptsize{($\pm$ 0.10)} \\
 \hline
\end{tabular}}
\label{tab:osr}
\end{table*}

\begin{table*}[t!]
\caption{Results for the PubFig dataset}
\centering
\resizebox{\textwidth}{!}{
\begin{tabular}{l|c|c|c|c|c|c|c|c|c|c|c|c}
 \textbf{Method} & \textbf{Male} & \textbf{White} & \textbf{Young} & \textbf{Smiling} & \textbf{Chubby} & \textbf{Forehead} & \textbf{Eyebrow} & \textbf{Eye} & \textbf{Nose} & \textbf{Lip} & \textbf{Face} & \textbf{Mean}  \\ \hline
 Relative Attributes~\cite{parikh2011} & 81.80 & 76.97 & 83.20 & 79.90 & 76.27 & 87.60 & 79.87 & 81.67 & 77.40 & 79.17 & 82.33 & 80.56 \\ 
 Relative Forest~\cite{Li2012RelativeFF} & 85.33 & 82.59 & 84.41 & 83.36 & 78.97 & 88.83 & 81.84 & 83.15 & 80.43 & 81.87 & 86.31 & 83.37 \\
 Fine-grained Comparison~\cite{Yu2014} & 91.77 & 87.43 & 91.87 & 87.00 & 87.37 & 94.00 & 89.83 & 91.40 & 89.07 & 90.43 & 86.70 & 89.72 \\
 VGG16-fc7 (baseline) & 85.56 & 80.59 & 85.20 & 84.81 & 82.56 & 88.50 & 83.50 & 83.11 & 81.52 & 85.67 & 86.23 & 84.30 \\
 \hline
 \multirow{2}{*}{RankNet (ours)} & 95.50 & 94.60 & 94.33 & 95.36 & 92.32 & 97.28 & 94.53 & 93.19 & 94.24 & 93.62 & 94.76 & \textbf{94.52} \\
                                 & \scriptsize{($\pm$ 0.36)} & \scriptsize{($\pm$ 0.55)} & \scriptsize{($\pm$ 0.36)} & \scriptsize{($\pm$ 0.56)} & \scriptsize{($\pm$ 0.36)} & \scriptsize{($\pm$ 0.49)} &  \scriptsize{($\pm$ 0.64)} & \scriptsize{($\pm$ 0.51)} & \scriptsize{($\pm$ 0.24)} & \scriptsize{($\pm$ 0.20)} & \scriptsize{($\pm$ 0.24)} & \scriptsize{($\pm$ 0.08)} \\
 \hline
\end{tabular}}
\label{tab:pubfig}
\end{table*}

Table \ref{tab:lfw} shows our results on the LFW-10 dataset. On this dataset, our method performs competitive with respect to the state-of-the-art, but cannot outperform it. We think this might be due to label noise in this dataset and due to the fact that most of the attributes in this dataset are highly local and methods that outperform us on this dataset look locally on regions of the image instead of the whole image.

\begin{table*}[t!]
\caption{Results for the LFW-10 dataset}
\centering
\resizebox{\textwidth}{!}{
\begin{tabular}{l|c|c|c|c|c|c|c|c|c|c|c}
 \textbf{Method} & \textbf{Bald} & \textbf{DkHair} & \textbf{Eyes} & \textbf{GdLook} & \textbf{Mascu.} & \textbf{Mouth} & \textbf{Smile} & \textbf{Teeth} & \textbf{FrHead} & \textbf{Young} & \textbf{Mean} \\ \hline
 Fine-grained Comparison~\cite{Li2012RelativeFF} & 67.9 & 73.6 & 49.6 & 64.7 & 70.1 & 53.4 & 59.7 & 53.5 & 65.6 & 66.2 & 62.4  \\
 Relative Attributes~\cite{parikh2011} & 70.4 & 75.7 & 52.6 & 68.4 & 71.3 & 55.0 & 54.6 & 56.0 & 64.5 & 65.8 & 63.4 \\
 Global + HOG~\cite{Verma2015ExploringLR} & 78.8 & 72.4 & 70.7 & 67.6 & 84.5 & 67.8 & 67.4 & 71.7 & 79.3 & 68.4 & 72.9 \\
 Relative Parts~\cite{Sandeep_2014_CVPR} & 71.8 & 80.5 & 90.5 & 77.6 & 67.0 & 77.6 & 81.3 & 76.2 & 80.2 & 82.4 & 78.5 \\
 Spatial Extent~\cite{spatial_extent} & 83.21 & 88.13 & 82.71 & 72.76 & 93.68 & 88.26 & 88.16 & 88.46 & 90.23 & 75.05 & \textbf{84.66} \\
 VGG16-fc7 (baseline) & 72.26 & 79.23 & 55.64 & 62.85 & 90.80 & 62.42 & 66.38 & 59.38 & 64.45 & 66.31 & 67.97 \\
 \hline
 \multirow{2}{*}{RankNet (ours)} & 81.14 & 88.92 & 74.44 & 70.28 & 98.08 & 85.46 & 82.49 & 82.77 & 81.90 & 76.33 & \textbf{82.18} \\ 
                                 & \scriptsize{($\pm$ 3.39)} & \scriptsize{($\pm$ 0.75)} & \scriptsize{($\pm$ 5.97)} & \scriptsize{($\pm$ 0.54)} & \scriptsize{($\pm$ 0.33)} & \scriptsize{($\pm$ 0.70)} & \scriptsize{($\pm$ 1.41)} & \scriptsize{($\pm$ 2.15)} & \scriptsize{($\pm$ 2.00)} & \scriptsize{($\pm$ 0.43)} & \scriptsize{($\pm$ 1.08)} \\
 \hline
\end{tabular}}
\label{tab:lfw}
\end{table*}

Tables \ref{tab:zap1} and \ref{tab:zap2} show the results on Zappos50K-1 and Zappos50K-2 datasets, respectively. Our method, again, achieves the state-of-the-art accuracy on both coarse-grained and fine-grained datasets. Our proposed method learns appropriate features for the task, given the large amount of training data available in this dataset.

\begin{table}[h]
\centering
\caption{Results for the UT-Zap50K-1 (coarse) dataset}
\resizebox{0.8\columnwidth}{!}{
\begin{tabular}{l|c|c|c|c|c}
 \textbf{Method} & \textbf{Open} & \textbf{Pointy} & \textbf{Sporty} & \textbf{Comfort} & \textbf{Mean} \\ \hline
 Relative Attributes~\cite{parikh2011} & 87.77 & 89.37 & 91.20 & 89.93 & 89.57 \\
 Fine-grained Comparison~\cite{Yu2014} & 90.67 & 90.83 & 92.67 & 92.37 & 91.64 \\
 Spatial Extent~\cite{spatial_extent} & 95.03 & 94.80 & 96.47 & 95.60 & 95.47 \\
 VGG16-fc7 (baseline) & 89.67 & 90.67 & 91.67 & 91.00 & 90.75 \\
 \hline
 \multirow{2}{*}{RankNet (ours)} & 95.37 & 94.43 & 97.30 & 95.57 & \textbf{95.67}\\
                                 & \scriptsize{($\pm$ 0.82)} & \scriptsize{($\pm$ 0.75)} & \scriptsize{($\pm$ 0.81)} & \scriptsize{($\pm$ 0.97)} & \scriptsize{($\pm$ 0.49)}\\
 \hline
\end{tabular}}
\label{tab:zap1}
\end{table}

\begin{table}[h]
\centering
\caption{Results for the UT-Zap50K-2 (fine-grained) dataset}
\resizebox{0.8\columnwidth}{!}{
\begin{tabular}{l|c|c|c|c|c}
 \textbf{Method} & \textbf{Open} & \textbf{Pointy} & \textbf{Sporty} & \textbf{Comfort} & \textbf{Mean} \\ \hline
 Relative Attributes~\cite{parikh2011} & 60.18 & 59.56 & 62.70 & 64.04 & 61.62  \\
 Fine-grained Comparison~\cite{Yu2014} & 74.91 & 63.74 & 64.54 & 62.51 & 66.43  \\
 LocalPair + ML + HOG~\cite{Verma2015ExploringLR} & 76.2 & 65.3 & 64.8 & 63.6 & 67.5 \\
 VGG16-fc7 (baseline) & 64.82 & 64.51 & 67.31 & 67.01 & 65.91 \\
 \hline
 \multirow{2}{*}{RankNet (ours)} & 73.45 & 68.20 & 73.07 & 70.31 & \textbf{71.26} \\
                                 & \scriptsize{($\pm$ 1.23)} & \scriptsize{($\pm$ 0.18)} & \scriptsize{($\pm$ 0.75)} & \scriptsize{($\pm$ 1.50)} & \scriptsize{($\pm$ 0.50)}\\
 \hline
\end{tabular}}
\label{tab:zap2}
\end{table}

\subsection{Qualitative Results} \label{sec.qres}

\begin{figure*}
\resizebox{\textwidth}{!}{
\centering
\includegraphics{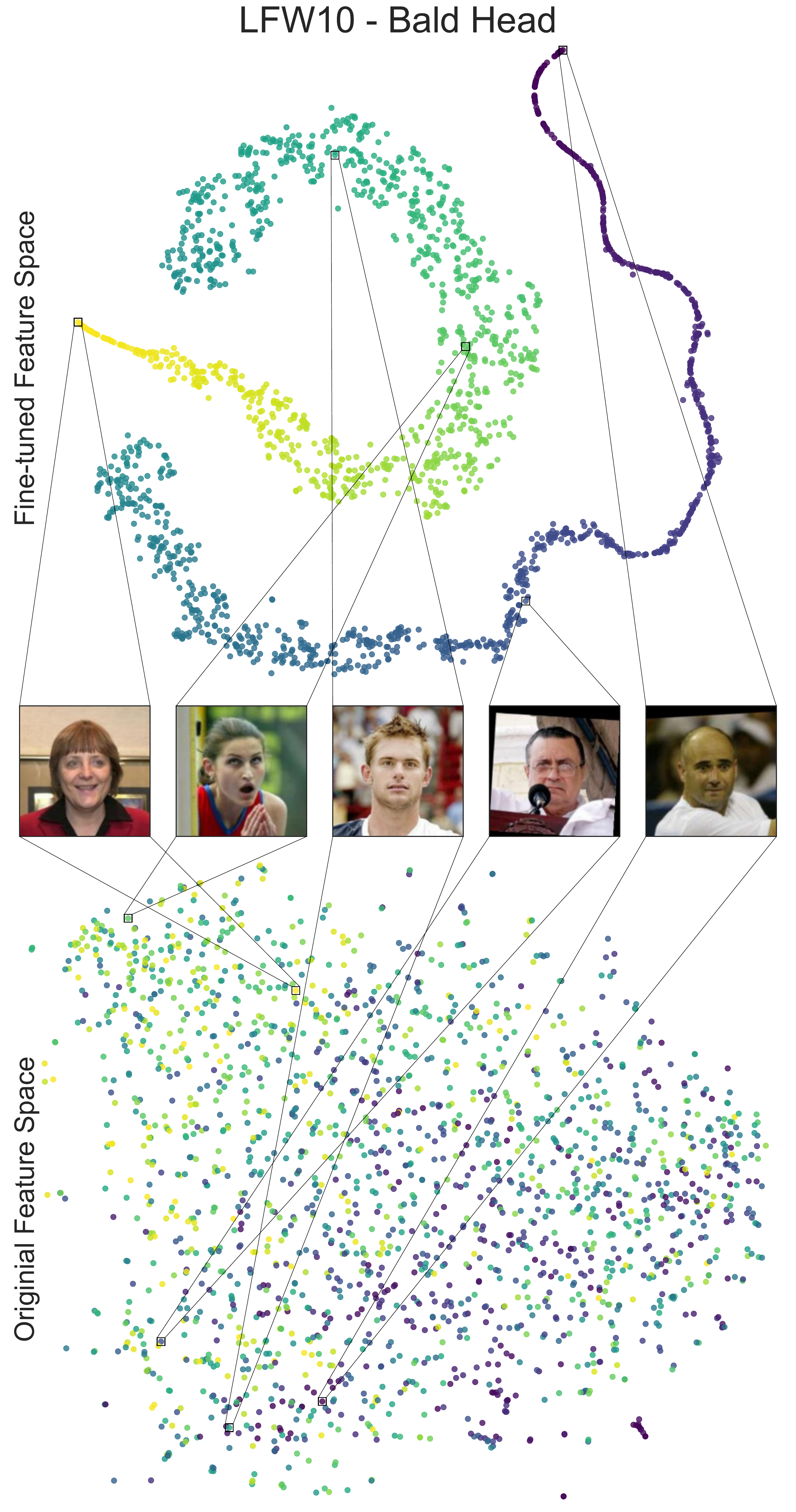}
\includegraphics{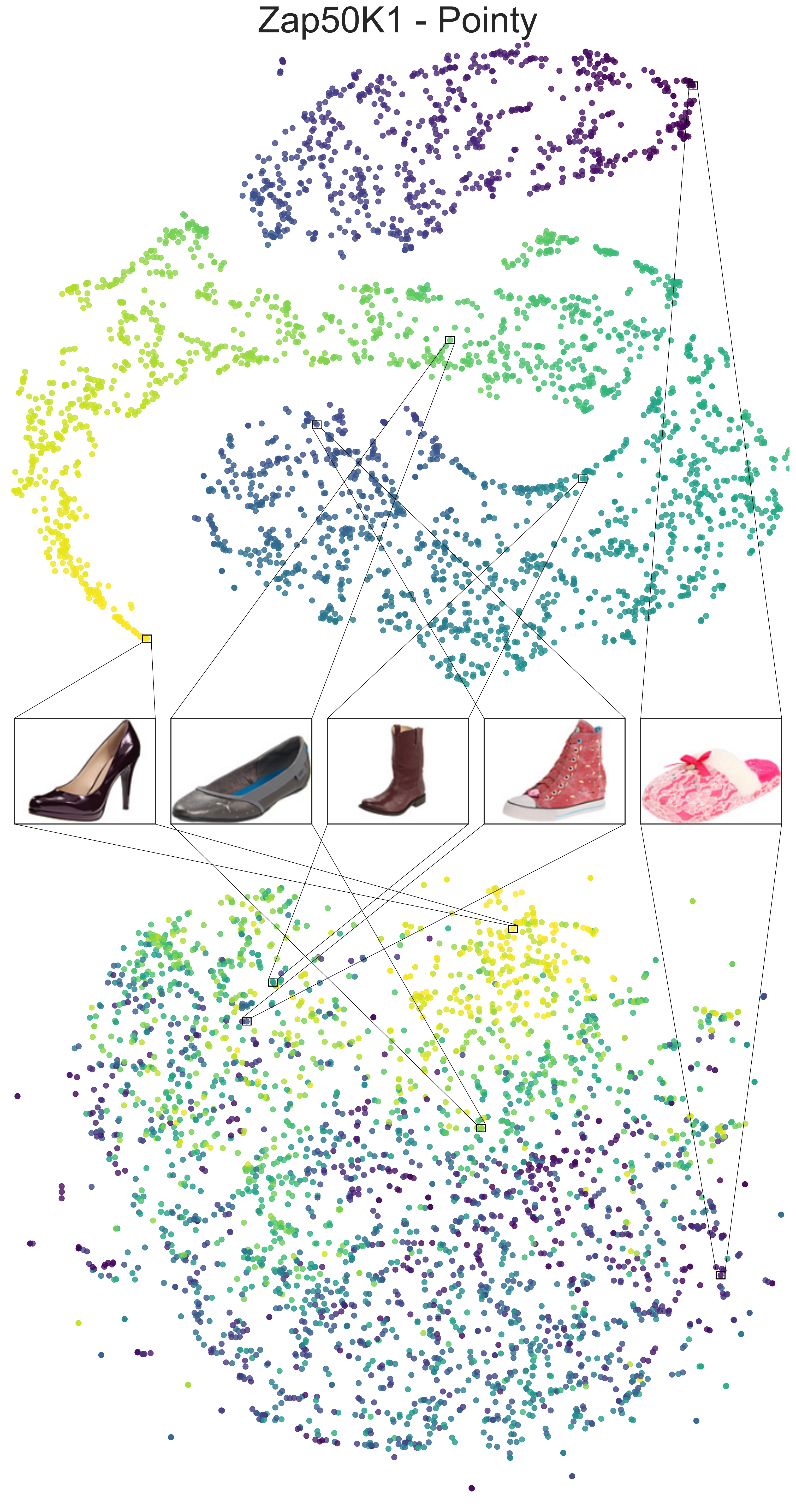}
}
\caption{t-SNE embedding of images in fine-tuned feature space (top) and original feature space (bottom). The set of visualizations on the left are for the \textit{Bald Head} attribute of the LFW-10 dataset, while the visualizations on the right are for the \textit{Pointy} attribute of the Zappos50K-1 dataset. Images in the middle row show a number of samples from the feature space. In the fine-tuned feature space, it is clear that images are ordered according to their value of the attribute. Each point is colored according to its value of the respective attribute, to discriminate images according to their value of the attribute.}
\label{featspace}
\end{figure*}

\begin{figure*}[t!]
\resizebox{\textwidth}{!}
{
    \begin{tikzpicture}
        \node [scale=3] (strong) {\textit{strong}}; 
        \node [scale=3, right=31cm of strong] (weak) {\textit{weak}};
        \path [draw, line width=3, <->] ([xshift=-2.7cm]strong.south) -- ([xshift=2cm]weak.south|-strong.south);

        \node [below=0.5cm of strong] (attr1im1) {\includegraphics[width=4cm, height=4cm]{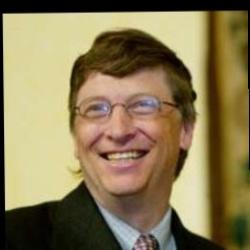}};
        \node (attr1name) [scale=3, left=0.5cm of attr1im1, align=center] {Smile \\ (LFW-10)};
        \node [right=0.5cm of attr1im1] (attr1im2) {\includegraphics[width=4cm, height=4cm]{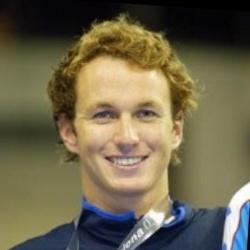}};
        \node [right=0.5cm of attr1im2] (attr1im3) {\includegraphics[width=4cm, height=4cm]{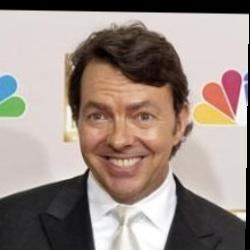}};
        \node [right=0.5cm of attr1im3] (attr1im4) {\includegraphics[width=4cm, height=4cm]{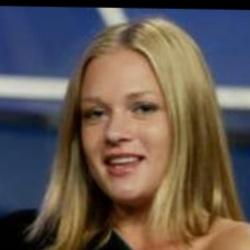}};
        \node [right=0.5cm of attr1im4] (attr1im5) {\includegraphics[width=4cm, height=4cm]{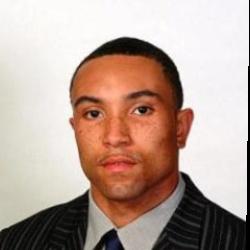}};
        \node [right=0.5cm of attr1im5] (attr1im6) {\includegraphics[width=4cm, height=4cm]{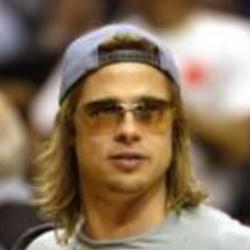}};
        \node [right=0.5cm of attr1im6] (attr1im7) {\includegraphics[width=4cm, height=4cm]{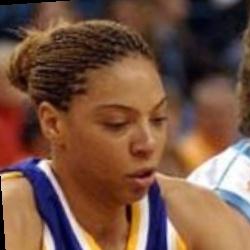}};
        \node [right=0.5cm of attr1im7] (attr1im8) {\includegraphics[width=4cm, height=4cm]{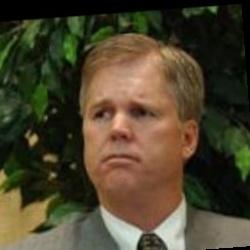}};

        \node [below=0.5cm of attr1im1] (attr4im1) {\includegraphics[width=4cm, height=3cm]{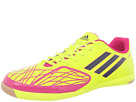}};
        \node (attr4name) [scale=3, below=1.8cm of attr1name, align=center] {Sporty \\ (Zap50K-1)};
        \node [right=0.5cm of attr4im1] (attr4im2) {\includegraphics[width=4cm, height=3cm]{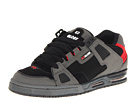}};
        \node [right=0.5cm of attr4im2] (attr4im3) {\includegraphics[width=4cm, height=3cm]{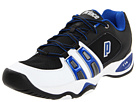}};
        \node [right=0.5cm of attr4im3] (attr4im4) {\includegraphics[width=4cm, height=3cm]{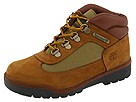}};
        \node [right=0.5cm of attr4im4] (attr4im5) {\includegraphics[width=4cm, height=3cm]{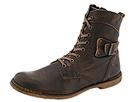}};
        \node [right=0.5cm of attr4im5] (attr4im6) {\includegraphics[width=4cm, height=3cm]{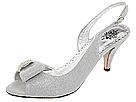}};
        \node [right=0.5cm of attr4im6] (attr4im7) {\includegraphics[width=4cm, height=3cm]{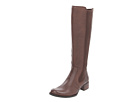}};
        \node [right=0.5cm of attr4im7] (attr4im8) {\includegraphics[width=4cm, height=3cm]{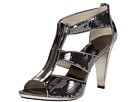}};

        \node [below=0.5cm of attr4im1] (attr5im1) {\includegraphics[width=4cm, height=4cm]{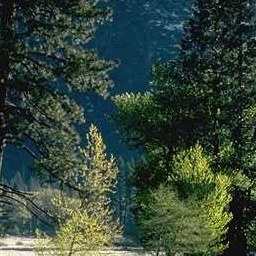}};
        \node (attr5name) [scale=3, below=1cm of attr4name, align=center] {Natural \\ (OSR)};
        \node [right=0.5cm of attr5im1] (attr5im2) {\includegraphics[width=4cm, height=4cm]{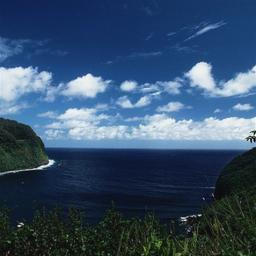}};
        \node [right=0.5cm of attr5im2] (attr5im3) {\includegraphics[width=4cm, height=4cm]{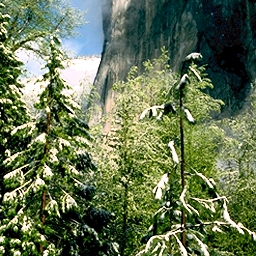}};
        \node [right=0.5cm of attr5im3] (attr5im4) {\includegraphics[width=4cm, height=4cm]{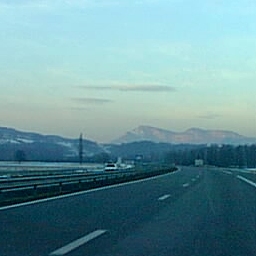}};
        \node [right=0.5cm of attr5im4] (attr5im5) {\includegraphics[width=4cm, height=4cm]{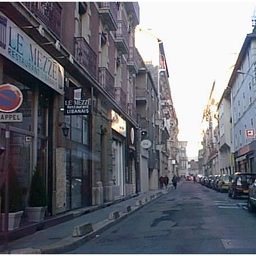}};
        \node [right=0.5cm of attr5im5] (attr5im6) {\includegraphics[width=4cm, height=4cm]{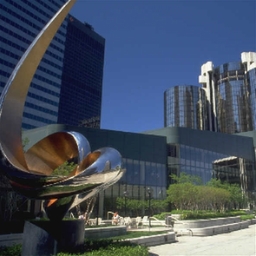}};
        \node [right=0.5cm of attr5im6] (attr5im7) {\includegraphics[width=4cm, height=4cm]{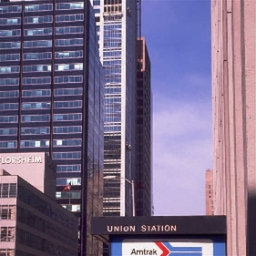}};
        \node [right=0.5cm of attr5im7] (attr5im8) {\includegraphics[width=4cm, height=4cm]{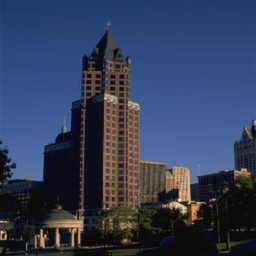}};

        \node [below=0.5cm of attr5im1] (attr7im1) {\includegraphics[width=4cm, height=4cm]{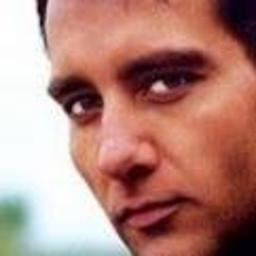}};
        \node (attr7name) [scale=3, below=1.8cm of attr5name, align=center] {Forehead \\ (PubFig)};
        \node [right=0.5cm of attr7im1] (attr7im2) {\includegraphics[width=4cm, height=4cm]{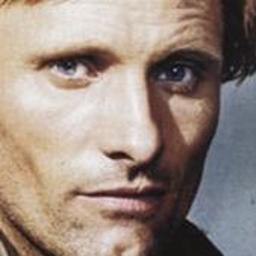}};
        \node [right=0.5cm of attr7im2] (attr7im3) {\includegraphics[width=4cm, height=4cm]{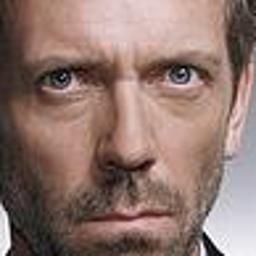}};
        \node [right=0.5cm of attr7im3] (attr7im4) {\includegraphics[width=4cm, height=4cm]{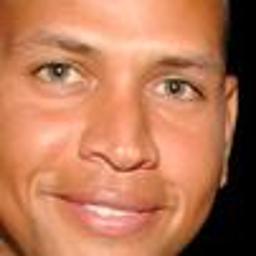}};
        \node [right=0.5cm of attr7im4] (attr7im5) {\includegraphics[width=4cm, height=4cm]{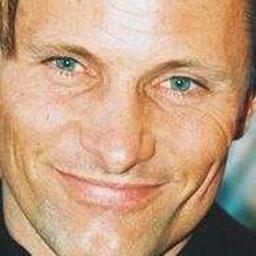}};
        \node [right=0.5cm of attr7im5] (attr7im6) {\includegraphics[width=4cm, height=4cm]{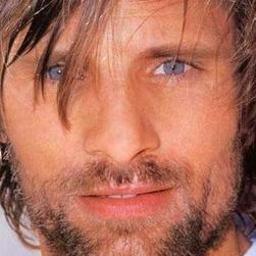}};
        \node [right=0.5cm of attr7im6] (attr7im7) {\includegraphics[width=4cm, height=4cm]{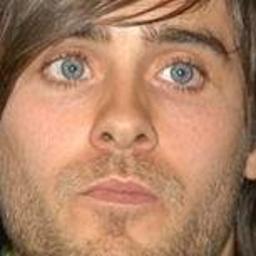}};
        \node [right=0.5cm of attr7im7] (attr7im8) {\includegraphics[width=4cm, height=4cm]{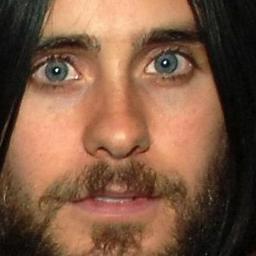}};

    \end{tikzpicture}
}
\caption{Sample images from different datasets, ordered according to the predicted value of their respective attribute.}
\label{figspectrum}
\end{figure*}

Our proposed method uses a deep network with two parts, the feature learning and extraction part and the ranking part. During training, not only the weights for the ranking part are learned, but also the weights for the feature learning and extraction part of the network, which were initialized using a pretrained network, are fine-tuned. By fine-tuning the features, our network learns a set of features that are more appropriate for the images of that particular dataset, along with the attribute of interest. To show the effectiveness of fine-tuning the features of the feature learning and extraction part of the network, we have projected them (features before and after fine-tuning) into 2-D space using the t-SNE \cite{van2008visualizing}, as can be seen in Figure \ref{featspace}. The visualizations on the top of each figure show the images projected into 2-D space from the fine-tuned feature space, while the visualizations on the bottom show the images from the original feature space. Each image is displayed as a point and colored according to its attribute strength. It is clear from these visualizations that fine-tuned feature space is better in capturing the ordering of images with respect to the respective attribute. Since t-SNE embedding is a non-linear embedding, relative distances between points in the high-dimensional space and the low-dimensional embedding space are preserved, thus close points in the low-dimensional embedding space are also close to each other in the high-dimensional space. It can, therefore, be seen that fine-tuning indeed changes the feature space such that images with similar values of the respective attribute get projected into a close vicinity of the feature space. However, in the original feature space, images are projected according to their visual content, regardless of their value of the attribute.

Another property of our network is that it can achieve a total ordering of images, given a set of pairwise orderings. In spite of the fact that training samples are pairs of images annotated according to their relative value of the attribute, the network can generalize the relativity of attribute values to a global ranking of images. Figure \ref{figspectrum} shows some images ordered according to their value of the respective attribute. 

\subsection{Saliency Maps and Localizing the Attributes} \label{sec.4.5}

\begin{figure*}
\captionsetup[subfigure]{labelformat=empty}
    \centering
    \begin{subfigure}
        \centering
        \includegraphics[width=0.8\linewidth, trim={0 1cm 0 0},clip]{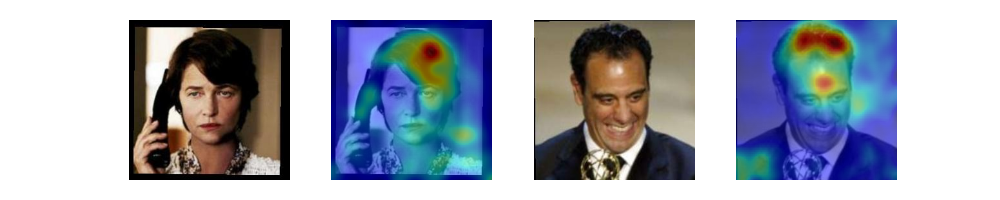}
        \footnotesize
        \stackunder[3pt]{
            \includegraphics[width=0.8\linewidth, trim={0 0 0 1cm},clip]{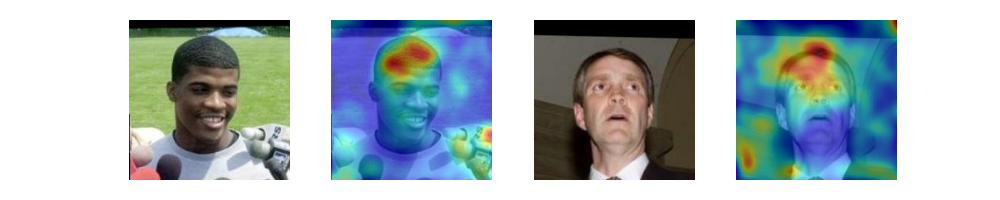}
        }{LFW10 - Bald Head}
        \vspace{0.4cm}
    \end{subfigure}
    
    \begin{subfigure}
        \centering
        \includegraphics[width=0.8\linewidth, trim={0 1cm 0 0},clip]{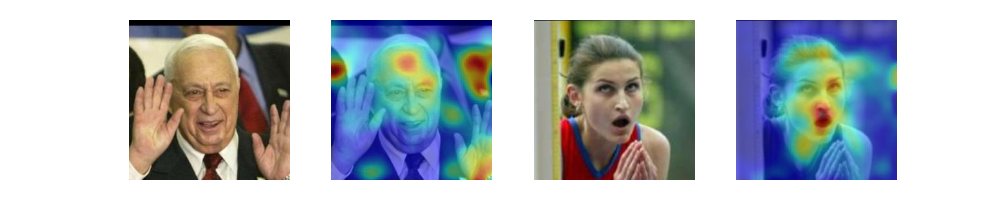}
        \footnotesize
        \stackunder[3pt]{
            \includegraphics[width=0.8\linewidth, trim={0 0 0 1cm},clip]{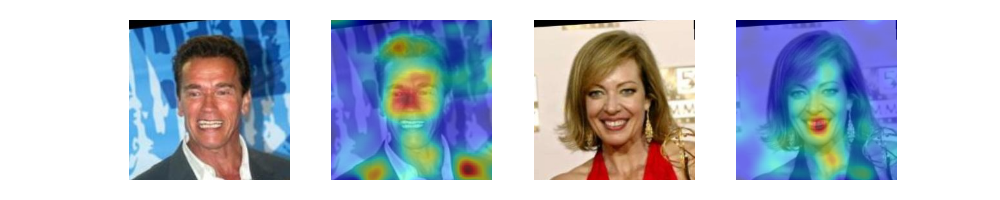}
        }{LFW10 - Good Looking}
        \vspace{0.4cm}
    \end{subfigure}
    
    \begin{subfigure}
        \centering
        \includegraphics[width=0.8\linewidth, trim={0 1cm 0 0},clip]{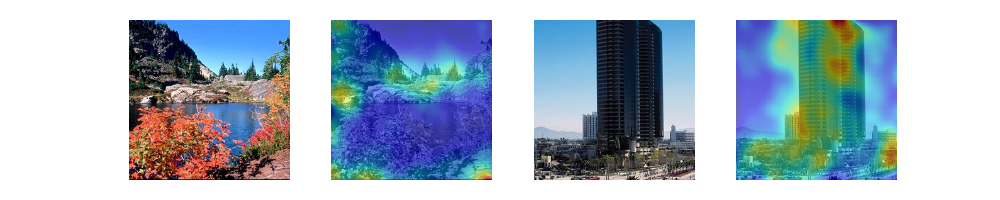}
        \footnotesize
        \stackunder[3pt]{
            \includegraphics[width=0.8\linewidth, trim={0 0 0 1cm},clip]{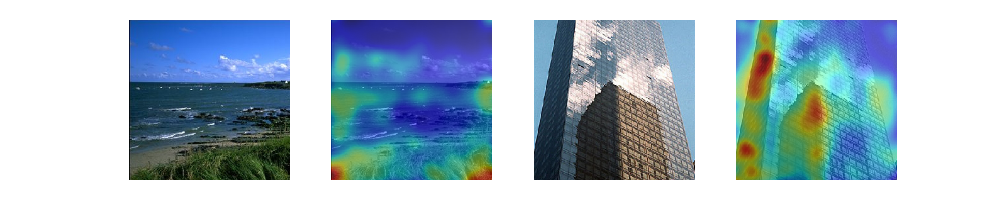}
        }{OSR - Natural}
        \vspace{0.4cm}
    \end{subfigure}
    
    \begin{subfigure}
        \centering
        \includegraphics[width=0.8\linewidth, trim={0 1cm 0 0},clip]{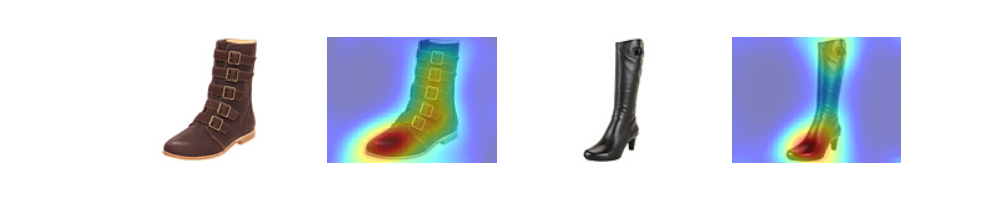}
        \footnotesize
        \stackunder[3pt]{
            \includegraphics[width=0.8\linewidth, trim={0 0 0 1cm} ,clip]{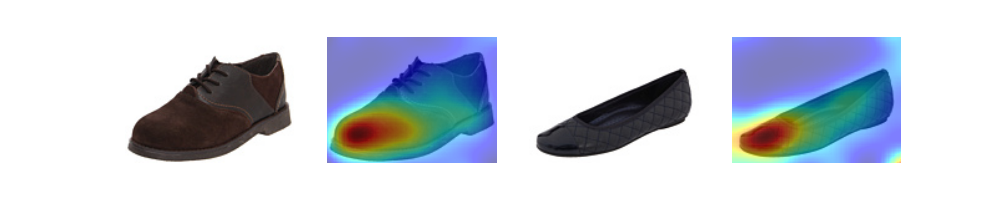}
        }{Zappos50k1 - Pointy}
    \end{subfigure}
    
    \caption{Saliency maps obtained from the network. First we feed two test images into the network and compute the derivative of the estimated posterior with respect to the pair of input images and use the method of \cite{saliency} to visualize salient pixels with Gaussian smoothing. In each row, the two input images from the a dataset's test set with their corresponding overlaid saliency maps are shown (the warmer the color of the overlay image, the more salient that pixel is).}
    \label{fig.5}
\end{figure*}

We have also used the method of \cite{saliency} to visualize the saliency of each attribute. Giving two image as inputs to the network, we take the derivative of the estimated posterior with respect to the input images and visualize them. Figure \ref{fig.5} shows some sample visualization for some test pairs. To generate this figure we have applied Gaussian smoothing to the saliency map.

These saliency maps visualize the pixels in the images which contributed most to the ranking predicted by the network. Sometimes these saliency maps are easily interpretable by humans and they can be used to localize attributes using the same network that was trained to rank the attributes in an unsupervised manner, \ie, although we haven't explicitly trained our network to localize the salient and informative regions of the image, it has implicitly learned to find these regions.
We see that this technique is able to localize both easy to localize attributes such as ``Bald Head" in the LFW10 dataset and abstract attributes such as ``Natural" in the OSR dataset.

\section{Conclusion}
\label{sec.5}

In this paper, we introduced an approach for relative attribute prediction on images, based on convolutional neural networks. Unlike previous methods that use engineered or hand-crafted features, our proposed method learns attribute-specific features, on-the-fly, during the learning procedure of the ranking function. Our results achieve state-of-the-art performance in relative attribute prediction on various datasets both coarse- and fine-grained.
We qualitatively show that the feature learning and extraction part, effectively learns appropriate features for each attribute and dataset.
Furthermore, we show that one can use a trained model for relative attribute prediction to obtain saliency maps for each attribute in the image.


\section{Acknowledgments}
We would like to thank Computer Engineering Department of Sharif University of Technology and HPC center of IPM for their support with computational resources.

\bibliographystyle{splncs}

\bibliography{refs}

\begin{thebibliography}{10}

\bibitem{whittlesearch}
Kovashka, A., Parikh, D., Grauman, K.:
\newblock Whittlesearch: {I}mage {S}earch with {R}elative {A}ttribute
  {F}eedback.
\newblock In: CVPR. (2012)

\bibitem{branson13}
Branson, S., Beijbom, O., Belongie, S.:
\newblock Efficient large-scale structured learning.
\newblock In: CVPR. (2013)

\bibitem{branson10}
Branson, S., Wah, C., Babenko, B., Schroff, F., Welinder, P., Perona, P.,
  Belongie, S.:
\newblock Visual recognition with humans in the loop.
\newblock In: ECCV. (2010)

\bibitem{6571196}
Lampert, C., Nickisch, H., Harmeling, S.:
\newblock Attribute-based classification for zero-shot visual object
  categorization.
\newblock IEEE TPAMI \textbf{36} (2014)  453--465

\bibitem{parikh2011}
Parikh, D., Grauman, K.:
\newblock Relative attributes.
\newblock CVPR (2011)  503--510

\bibitem{ferrari2007learning}
Ferrari, V., Zisserman, A.:
\newblock Learning visual attributes.
\newblock In: NIPS. (2007)  433--440

\bibitem{Farhadi09describingobjects}
Farhadi, A., Endres, I., Hoiem, D., Forsyth, D.:
\newblock Describing objects by their attributes.
\newblock In: CVPR. (2009)

\bibitem{Aude01}
Oliva, A., Torralba, A.:
\newblock Modeling the shape of the scene: A holistic representation of the
  spatial envelope.
\newblock IJCV \textbf{42} (2001)  145--175

\bibitem{hog}
Dalal, N., Triggs, B.:
\newblock Histograms of oriented gradients for human detection.
\newblock In: CVPR. (2005)  886--893

\bibitem{Krizhevsky2012ImageNetCW}
Krizhevsky, A., Sutskever, I., Hinton, G.E.:
\newblock Imagenet classification with deep convolutional neural networks.
\newblock In: NIPS. (2012)

\bibitem{RCNN}
Girshick, R., Donahue, J., Darrell, T., Malik, J.:
\newblock Rich feature hierarchies for accurate object detection and semantic
  segmentation.
\newblock In: CVPR. (2014)

\bibitem{fullyconv}
Long, J., Shelhamer, E., Darrell, T.:
\newblock Fully convolutional networks for semantic segmentation.
\newblock In: CVPR. (2015)  3431--3440

\bibitem{offtheshelf}
Razavian, A.S., Azizpour, H., Sullivan, J., Carlsson, S.:
\newblock Cnn features off-the-shelf: an astounding baseline for recognition.
\newblock In: CVPRW. (2014)  512--519

\bibitem{saliency}
Simonyan, K., Vedaldi, A., Zisserman, A.:
\newblock Deep inside convolutional networks: Visualising image classification
  models and saliency maps.
\newblock arXiv preprint arXiv:1312.6034 (2013)

\bibitem{Farhadi2010EveryPT}
Farhadi, A., Hejrati, M., Sadeghi, M.A., Young, P., Rashtchian, C.,
  Hockenmaier, J., Forsyth, D.A.:
\newblock Every picture tells a story: Generating sentences from images.
\newblock In: ECCV. (2010)

\bibitem{7298613}
Tao, R., Smeulders, A.W., Chang, S.F.:
\newblock Attributes and categories for generic instance search from one
  example.
\newblock In: CVPR. (2015)  177--186

\bibitem{6838985}
Khan, F., van~de Weijer, J., Anwer, R., Felsberg, M., Gatta, C.:
\newblock Semantic pyramids for gender and action recognition.
\newblock IEEE TIP \textbf{23} (2014)  3633--3645

\bibitem{5995353}
Liu, J., Kuipers, B., Savarese, S.:
\newblock Recognizing human actions by attributes.
\newblock In: CVPR. (2011)  3337--3344

\bibitem{6475038}
Liu, J., Yu, Q., Javed, O., Ali, S., Tamrakar, A., Divakaran, A., Cheng, H.,
  Sawhney, H.:
\newblock Video event recognition using concept attributes.
\newblock In: WACV. (2013)  339--346

\bibitem{KovashkaG13}
Kovashka, A., Grauman, K.:
\newblock Attribute pivots for guiding relevance feedback in image search.
\newblock In: ICCV. (2013)  297--304

\bibitem{Joachims2002}
Joachims, T.:
\newblock Optimizing search engines using clickthrough data.
\newblock In: ACM KDD. (2002)  133--142

\bibitem{Li2012RelativeFF}
Li, S., Shan, S., Chen, X.:
\newblock Relative forest for attribute prediction.
\newblock In: ACCV. (2012)

\bibitem{5771429}
Datta, A., Feris, R., Vaquero, D.:
\newblock Hierarchical ranking of facial attributes.
\newblock In: FG. (2011)  36--42

\bibitem{6909607}
Jayaraman, D., Sha, F., Grauman, K.:
\newblock Decorrelating semantic visual attributes by resisting the urge to
  share.
\newblock In: CVPR. (2014)  1629--1636

\bibitem{1641014}
Zhang, H., Berg, A., Maire, M., Malik, J.:
\newblock {SVM-KNN}: Discriminative nearest neighbor classification for visual
  category recognition.
\newblock In: CVPR. Volume~2. (2006)  2126--2136

\bibitem{Yu2014}
Yu, A., Grauman, K.:
\newblock Fine-grained visual comparisons with local learning.
\newblock In: CVPR. (2014)

\bibitem{Yu2015}
Yu, A., Grauman, K.:
\newblock Just noticeable differences in visual attributes.
\newblock In: ICCV. (2015)

\bibitem{LeCun1989HandwrittenDR}
LeCun, Y., Boser, B.E., Denker, J.S., Henderson, D., Howard, R.E., Hubbard,
  W.E., Jackel, L.D.:
\newblock Handwritten digit recognition with a back-propagation network.
\newblock In: NIPS. (1989)

\bibitem{6909475}
Girshick, R., Donahue, J., Darrell, T., Malik, J.:
\newblock Rich feature hierarchies for accurate object detection and semantic
  segmentation.
\newblock In: CVPR. (2014)  580--587

\bibitem{6909608}
Zhang, N., Paluri, M., Ranzato, M., Darrell, T., Bourdev, L.:
\newblock {PANDA}: Pose aligned networks for deep attribute modeling.
\newblock In: CVPR. (2014)  1637--1644

\bibitem{Escorcia_2015_CVPR}
Escorcia, V., Carlos~Niebles, J., Ghanem, B.:
\newblock On the relationship between visual attributes and convolutional
  networks.
\newblock In: CVPR. (2015)

\bibitem{Shankar_2015_CVPR}
Shankar, S., Garg, V.K., Cipolla, R.:
\newblock Deep-carving: Discovering visual attributes by carving deep neural
  nets.
\newblock In: CVPR. (2015)

\bibitem{khan15}
Khan, F.S., Anwer, R.M., van~de Weijer, J., Felsberg, M., Laaksonen, J.:
\newblock Deep semantic pyramids for human attributes and action recognition.
\newblock In: Image Analysis.
\newblock Springer (2015)  341--353

\bibitem{Huang_2015_ICCV}
Huang, J., Feris, R.S., Chen, Q., Yan, S.:
\newblock Cross-domain image retrieval with a dual attribute-aware ranking
  network.
\newblock In: ICCV. (2015)

\bibitem{Burges2005}
Burges, C., Shaked, T., Renshaw, E., Lazier, A., Deeds, M., Hamilton, N.,
  Hullender, G.:
\newblock Learning to rank using gradient descent.
\newblock In: ICML. (2005)  89--96

\bibitem{song2014}
Song, Y., Wang, H., He, X.:
\newblock Adapting deep ranknet for personalized search.
\newblock In: WSDM. (2014)

\bibitem{Wan2014}
Wan, J., Wang, D., Hoi, S.C.H., Wu, P., Zhu, J., Zhang, Y., Li, J.:
\newblock Deep learning for content-based image retrieval: A comprehensive
  study.
\newblock In: ACM MM. (2014)  157--166

\bibitem{YaoCVPR2016}
Yao, T., Mei, T., Rui, Y.:
\newblock Highlight detection with pairwise deep ranking for first-person video
  summarization.
\newblock In: CVPR. (2016)

\bibitem{verydeep}
Simonyan, K., Zisserman, A.:
\newblock Very deep convolutional networks for large-scale image recognition.
\newblock arXiv preprint arXiv:1409.1556 (2014)

\bibitem{googlenet}
Szegedy, C., Liu, W., Jia, Y., Sermanet, P., Reed, S., Anguelov, D., Erhan, D.,
  Vanhoucke, V., Rabinovich, A.:
\newblock Going deeper with convolutions.
\newblock In: CVPR. (2015)

\bibitem{Sandeep_2014_CVPR}
Sandeep, R.N., Verma, Y., Jawahar, C.V.:
\newblock Relative parts: Distinctive parts for learning relative attributes.
\newblock In: CVPR. (2014)

\bibitem{lasagne}
Dieleman, S., Schlüter, J., Raffel, C., Olson, E., Sønderby, S.K., Nouri, D.,
  Maturana, D., Thoma, M., Battenberg, E., Kelly, J., Fauw, J.D., Heilman, M.,
  diogo149, McFee, B., Weideman, H., takacsg84, peterderivaz, Jon, instagibbs,
  Rasul, D.K., CongLiu, Britefury, Degrave, J.:
\newblock Lasagne: First release. (2015)

\bibitem{ilsvrc2014}
Russakovsky, O., Deng, J., Su, H., Krause, J., Satheesh, S., Ma, S., Huang, Z.,
  Karpathy, A., Khosla, A., Bernstein, M.,  et~al.:
\newblock Imagenet large scale visual recognition challenge.
\newblock International Journal of Computer Vision \textbf{115} (2015)
  211--252

\bibitem{glorot}
Glorot, X., Bengio, Y.:
\newblock Understanding the difficulty of training deep feedforward neural
  networks.
\newblock In: AISTATS. (2010)  249--256

\bibitem{Tieleman2012}
Tieleman, T., Hinton, G.:
\newblock {Lecture 6.5---RmsProp: Divide the gradient by a running average of
  its recent magnitude}.
\newblock COURSERA: Neural Networks for Machine Learning (2012)

\bibitem{Verma2015ExploringLR}
Verma, Y., Jawahar, C.V.:
\newblock Exploring locally rigid discriminative patches for learning relative
  attributes.
\newblock In: BMVC. (2015)

\bibitem{spatial_extent}
Xiao, F., Jae~Lee, Y.:
\newblock Discovering the spatial extent of relative attributes.
\newblock In: CVPR. (2015)

\bibitem{van2008visualizing}
Van~der Maaten, L., Hinton, G.:
\newblock Visualizing data using {t-SNE}.
\newblock JMLR \textbf{9} (2008) ~85

\end{thebibliography}



\end{document}